\pdfoutput=1

\documentclass[11pt]{article}
\usepackage{acl}

\usepackage{times}
\usepackage{latexsym}

\usepackage[T1]{fontenc}

\usepackage[utf8]{inputenc}

\usepackage{microtype}

\usepackage{inconsolata}

 \usepackage{graphicx} 
\usepackage{natbib}  
\usepackage{caption} 
\usepackage{algorithm}
\usepackage{algorithmic}
\usepackage{microtype}
\usepackage{comment}
\usepackage{todonotes}
\usepackage{color,soul}
\usepackage{graphicx}
\usepackage{pbox}
\usepackage{subcaption}
\usepackage{epstopdf}
\usepackage{xstring}
\usepackage{fdsymbol}
\usepackage{multirow}
\usepackage{mathtools}
\usepackage[normalem]{ulem}
\usepackage{pifont}
\usepackage{todonotes}
\usepackage{tikz-dependency}
\usepackage{diagbox}
\usepackage{array}
\usepackage{lipsum}
\usepackage[normalem]{ulem}
\usepackage{booktabs}
\useunder{\uline}{\ul}{}
\usepackage{enumitem,kantlipsum}
\usepackage{mathtools}

\usepackage{hhline}
\usepackage{multirow}
\usepackage{ulem}
\usepackage{newfloat}
\usepackage{listings}
\DeclareCaptionStyle{ruled}{labelfont=normalfont,labelsep=colon,strut=off} 
\floatstyle{ruled}
\newfloat{listing}{tb}{lst}{}
\floatname{listing}{Listing}
\usepackage{amsmath,amsfonts,bm}

\def\eqref#1{equation~\ref{#1}}

\def\1{\bm{1}}

\DeclareMathAlphabet{\mathsfit}{\encodingdefault}{\sfdefault}{m}{sl}
\SetMathAlphabet{\mathsfit}{bold}{\encodingdefault}{\sfdefault}{bx}{n}

\usepackage{comment}

\usepackage{todonotes}

\title{Scaling up Discovery of Latent Concepts in Deep NLP Models}

\author{
Majd Hawasly \hspace{11mm} Fahim Dalvi  \hspace{11mm} Nadir Durrani \\
Qatar Computing Research Institute, HBKU, Doha, Qatar \\ 
{\tt \{mhawasly,faimaduddin,ndurrani\}@hbku.edu.qa} \\ 
}

\begin{document}

\maketitle

\begin{abstract}

Despite the revolution caused by deep NLP models, they remain black boxes, necessitating research to understand their decision-making processes. A recent work by \citet{dalvi2022discovering} carried out representation analysis through the lens of clustering latent spaces within pre-trained models (PLMs), but that approach is limited to small scale due to the high cost of running Agglomerative hierarchical clustering. This paper studies clustering algorithms in order to scale the discovery of encoded concepts in PLM representations to larger datasets and models. We propose metrics for assessing the quality of discovered latent concepts and use them to compare the studied clustering algorithms. We found that K-Means-based concept discovery significantly enhances efficiency while maintaining the quality of the obtained concepts. 
Furthermore, we demonstrate the practicality of this newfound efficiency by scaling latent concept discovery to LLMs and phrasal concepts.\footnote{Source Code: \url{https://github.com/qcri/Latent_Concept_Analysis}}

\end{abstract}
\section{Introduction}

Transformer-based language models excel at revealing intricate patterns, semantic relationships, and nuanced linguistic dependencies concealed within vast textual datasets through unsupervised learning. Their capability to encode complex abstractions, surpassing surface-level word meanings, has resulted in significant advancements across various natural language understanding tasks. A considerable body of research dedicated to interpreting pre-trained language models (e.g. \citet{belinkov:2017:acl,tenney-etal-2019-bert,geva-etal-2021-transformer,sajjad-etal-2022-neuron} among others) seeks to answer the question: \emph{What knowledge is learned within these models?} Researchers have delved into the concepts encoded in pre-trained language models by probing them against various linguistic properties. Our work facilitates this line of work in interpretability by scaling up discovery of latent concepts learned within pre-trained language models.

\citet{Mikolov:2013:ICLR} demonstrated that words exhibit a tendency to form clusters in high-dimensional spaces, reflecting their morphological, syntactic, and semantic relationships. Building upon this foundational insight, recent studies \citep{michael-etal-2020-asking, dalvi2022discovering, Yao-Latent} delve into representation analysis by exploring latent spaces within pre-trained models. \citet{dalvi2022discovering} discovered encoded concepts in pre-trained models by employing Agglomerative hierarchical clustering \citep{gowda1978Agglomerative} on the contextualized representations in the BERT model \citep{devlin-etal-2019-bert}. However, a fundamental limitation of their work 
is the computational expense of the underlying methodology. Since contextual representations are high-dimensional, only a limited amount of data can be clustered to extract the latent concepts. This significantly undermines the purpose of concept discovery, providing only a limited perspective on the spectrum of concepts that might be learned within the model and significantly limiting the scalability of the approach. 
	
	\begin{figure*}[t]
		\centering
		\includegraphics[width=0.80\linewidth]{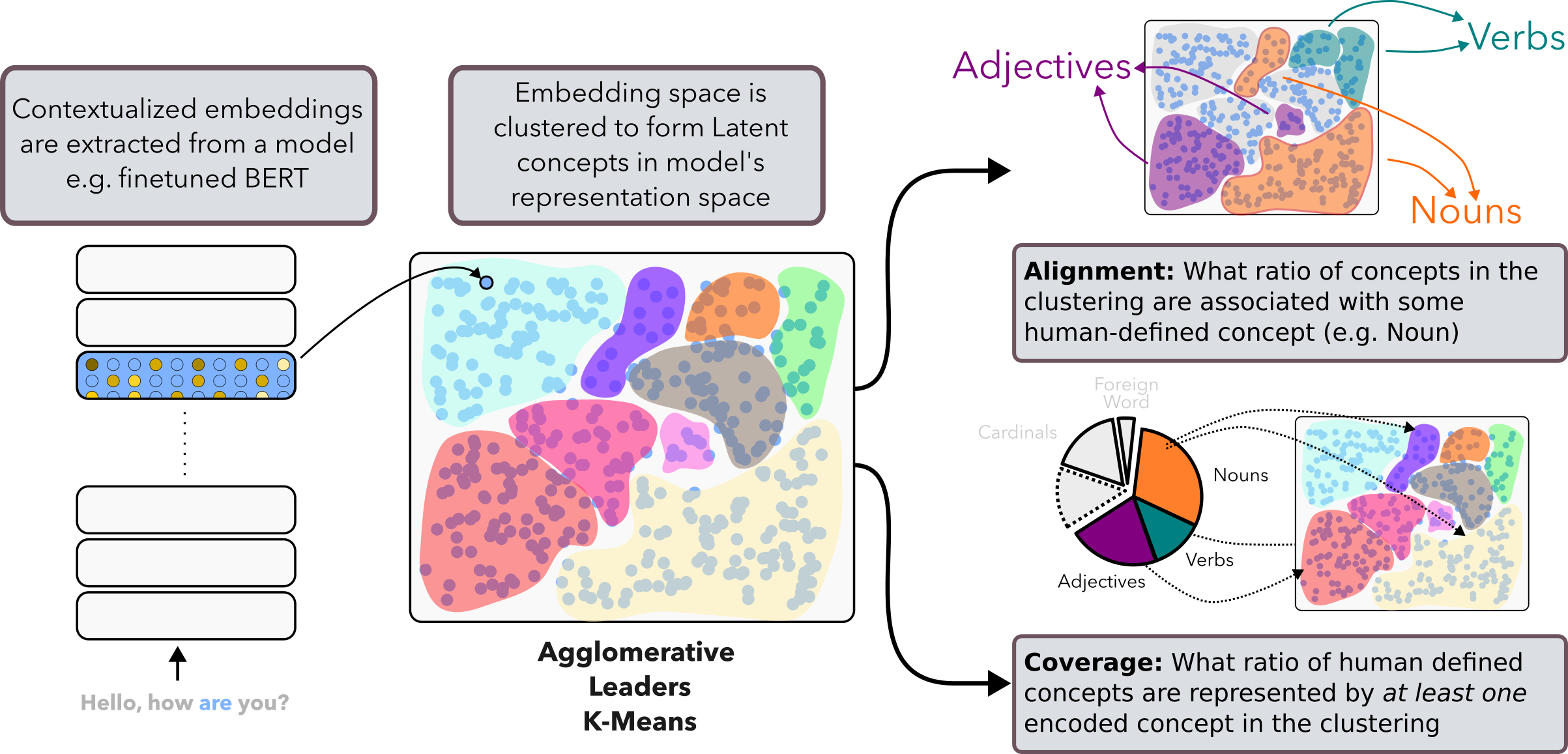}
		
		\caption{Discovery of encoded concepts within a PLM using clustering of contextualized embeddings, and evaluation of discovered concepts through \textit{alignment} and \textit{coverage} metrics with respect to human ontologies.
		}
		\label{fig:pipeline}
	\end{figure*}
	
	In this work, we aim to address this 
 shortcoming by employing computationally-cheaper clustering algorithms, specifically comparing three algorithms in quality and computational efficiency: \emph{Agglomerative Hierarchical Clustering}, \emph{Leaders Algorithm}, and \emph{$K$-Means Clustering}. As there is no inherent groundtruth clustering we can rely on to measure the quality of the algorithms in the latent space, we introduce a metric with two dimensions, \emph{alignment} and \emph{coverage}, to measure the "goodness" of a clustering. We also show that scaling the underlying data for concept discovery results in significantly better results, as well as enables new directions that were previously unexplored. In summary we make the following contributions:
	
	\begin{itemize}
		\item We present a comprehensive comparison between various clustering techniques regarding their quality and efficiency for the task of latent concept discovery.
		\item We introduce a metric with two dimensions to measure the quality of extracted latent concepts: \textit{alignment} and \textit{coverage} of linguistic ontologies.
		\item We demonstrate that K-Means exhibits the capacity to handle vast datasets effectively  while still producing latent concepts of roughly the same quality as Agglomerative hierarchical clustering.
		\item We show that increasing the size of the dataset used for clustering leads to higher-quality concept discovery, improving the coverage of POS tags by 8\% on average in the last layer of fine-tuned BERT models, and 26\% in the base Llama2 model.
		\item We present preliminary results in two new directions that K-Means affords us: exploration of latent concepts at a level higher than just words (e.g. phrases), and scaling concept discovery to large language models (LLMs).
	\end{itemize}

\section{Concept Discovery}
\label{sec:conceptDiscovery}

Our investigation builds upon the work of discovering Latent Ontologies in contextualized representations \cite{dalvi2022discovering}. At a high level, feature vectors (contextualized representations) are initially generated by performing a forward pass on a pre-trained language model. The representations are then clustered to uncover the encoded concepts of the model (See Figure~\ref{fig:pipeline} for illustration). A concept, in this context, can be understood as a collection of words grouped together based on some linguistic relationship, such as lexical, semantic, syntactic, or morphological connections. Figure~\ref{fig:aligned_concept_examples} showcases concepts within the latent space of the BERT model, wherein word representations are arranged based on distinct linguistic concepts.

Formally, consider a pre-trained model $\mathbf{M}$ with $L$ layers: $l_1, l_2, \ldots, l_L$. Using a dataset of $S$ sentences totaling $N$ tokens, $\mathcal{D}=[w_1, w_2, \ldots, w_N]$, we generate feature vectors: $\mathcal{D} \xrightarrow{\mathbf{M}_l} \mathbf{z}^l = [\mathbf{z}^l_1, \ldots, \mathbf{z}^l_N]$, where $\mathbf{z}_i^l$ is the contextualized representation for the word $w_i$ within the context of its sentence at layer $l$. A clustering algorithm is then employed in the per-layer feature vector space $\mathbf{z}^l$ to  discover layer-$l$ encoded concepts.

\subsection{Clustering algorithms}
In this paper, our focus is to increase the scalability of latent concept discovery. Hence, we evaluate different clustering algorithms in order to find one that can produce similar or better categorization of the latent space of a model with higher computational efficiency than the originally proposed method. To this end, we study three algorithms:

\begin{figure*}[t]
    \begin{subfigure}[b]{0.21\linewidth}
    \centering
    \includegraphics[height=2.7cm]{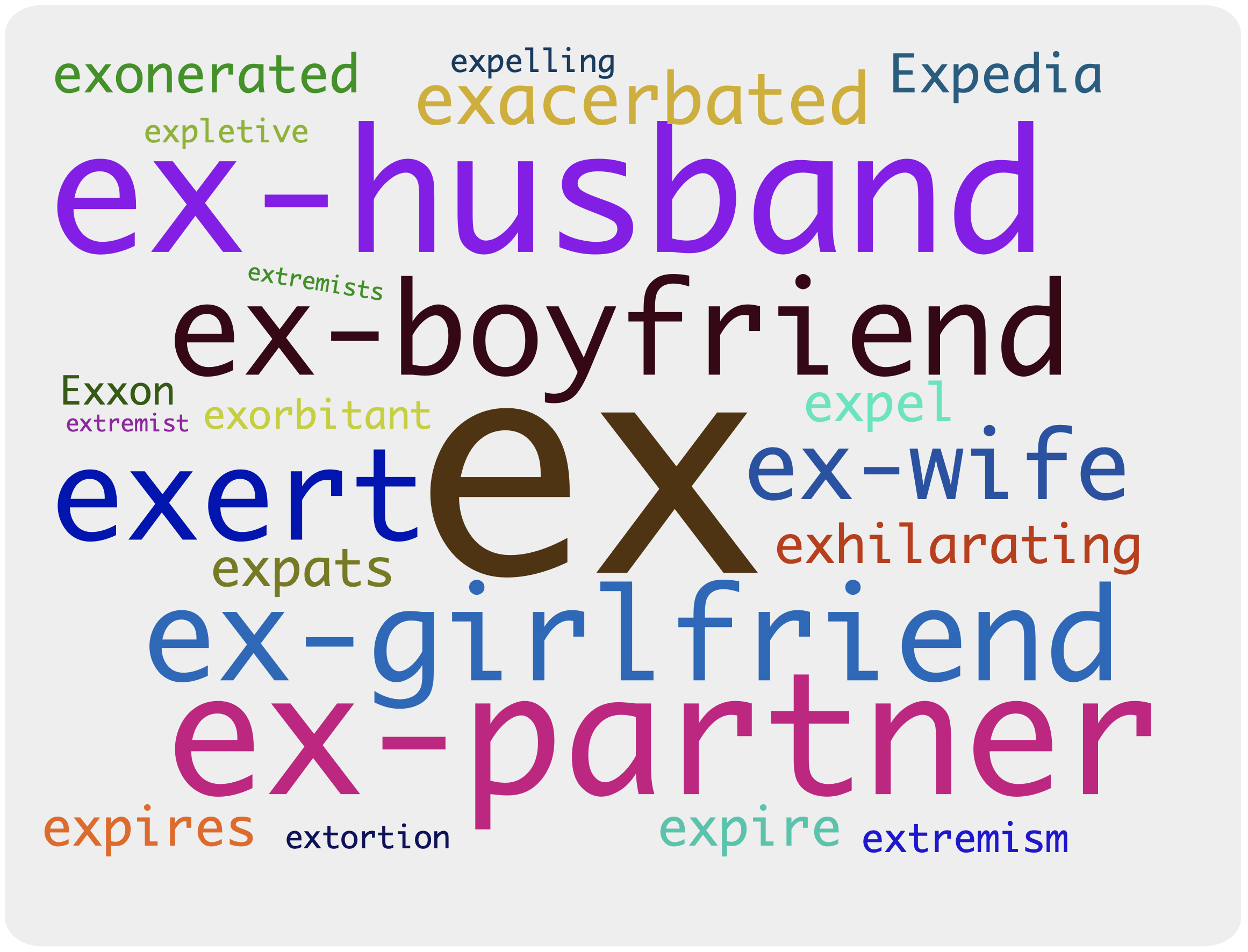}
    \caption{Lexical: \textit{ex-}}
    \label{fig:ngram}
    \end{subfigure} \hspace{5mm}
    \centering
    \begin{subfigure}[b]{0.23\linewidth}
    \centering
     \includegraphics[height=2.7cm]{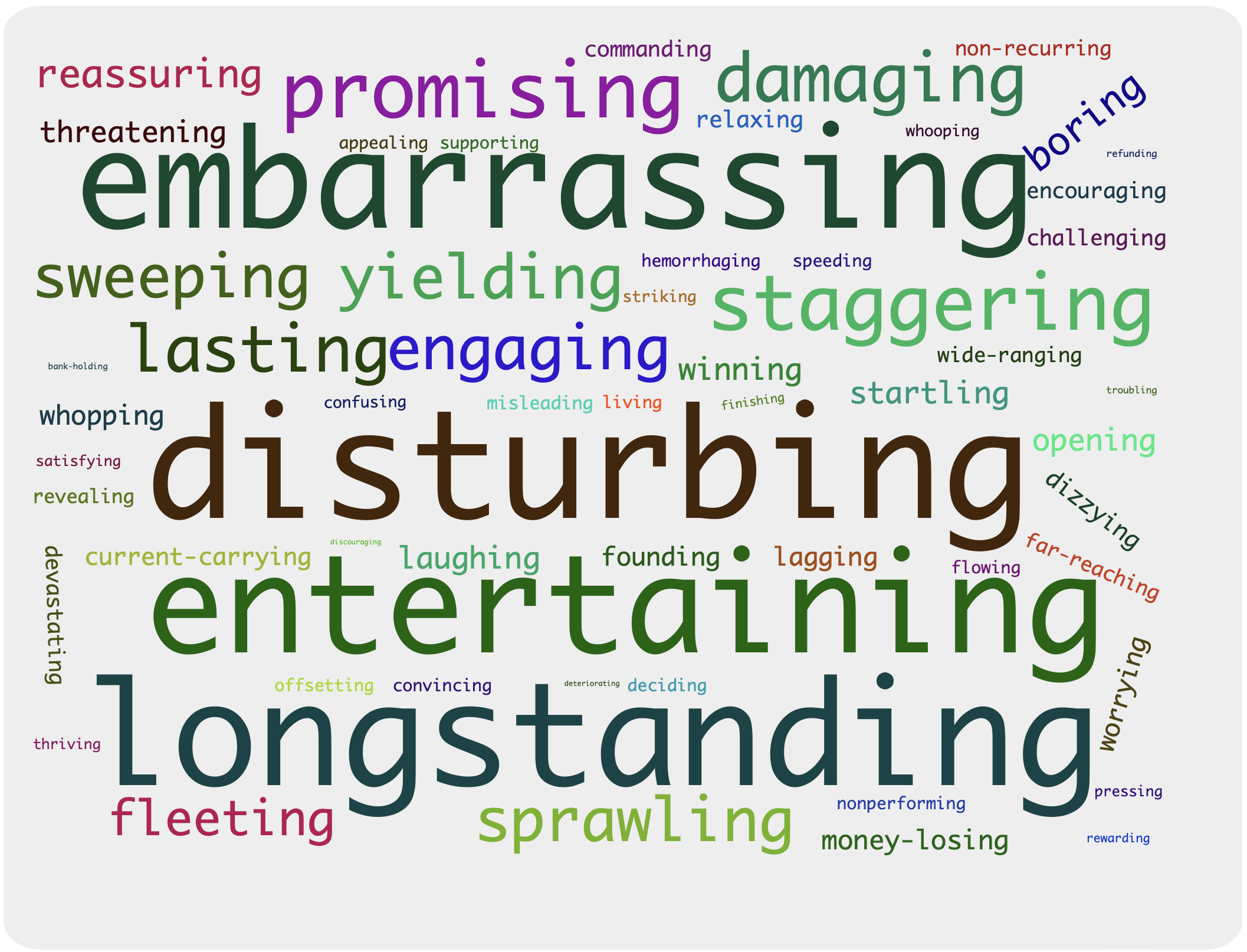}
    \caption{Morphological: Adjective}
    \label{fig:JJ+NN}
    \end{subfigure} \hspace{5mm}
    \centering
    \begin{subfigure}[b]{0.21\linewidth}
    \centering
    \includegraphics[height=2.7cm]{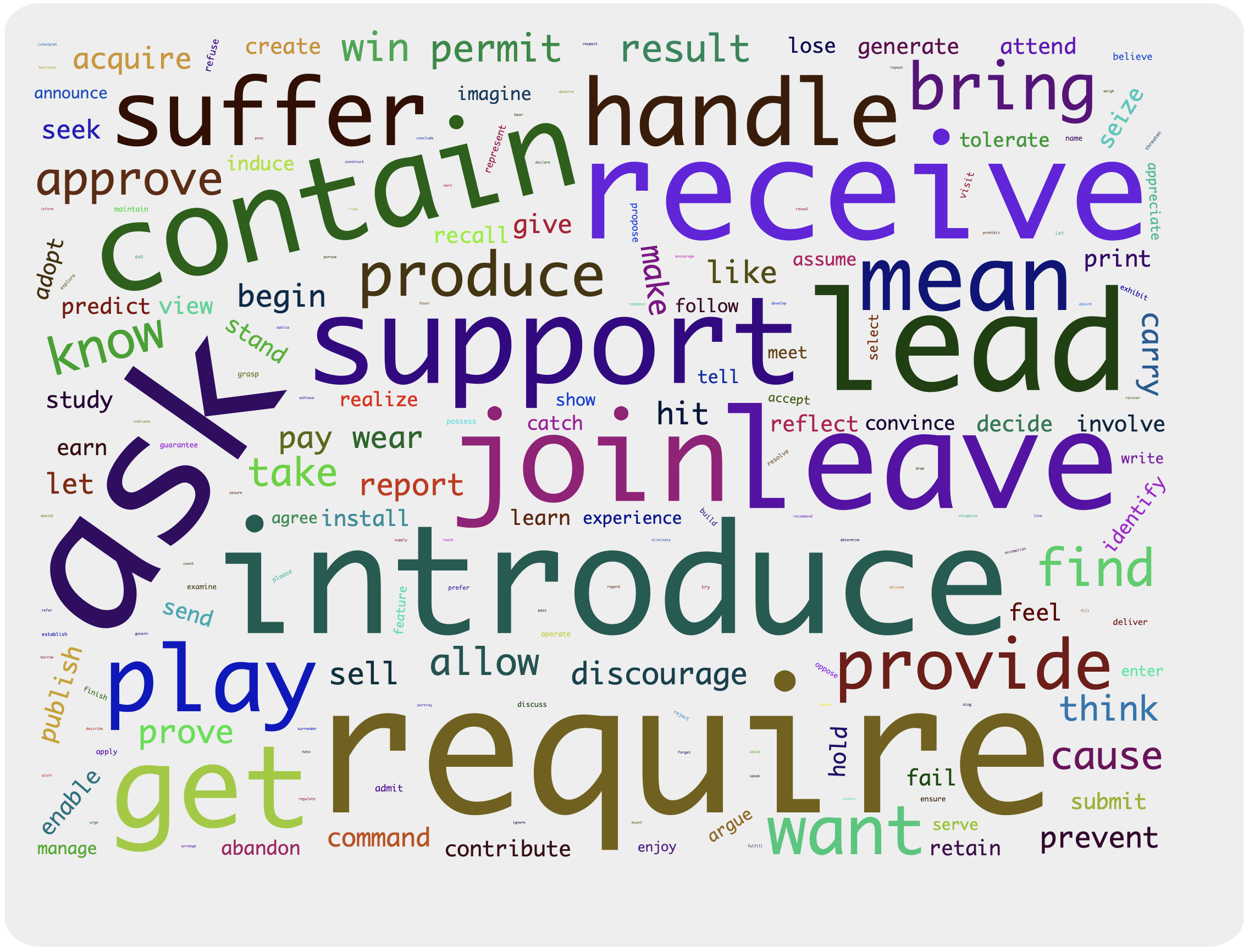}
    \caption{Syntactic: Verb}
    \label{fig:ccg+V}
    \end{subfigure} \hspace{5mm}
    \begin{subfigure}[b]{0.21\linewidth}
    \centering
   \includegraphics[height=2.7cm]{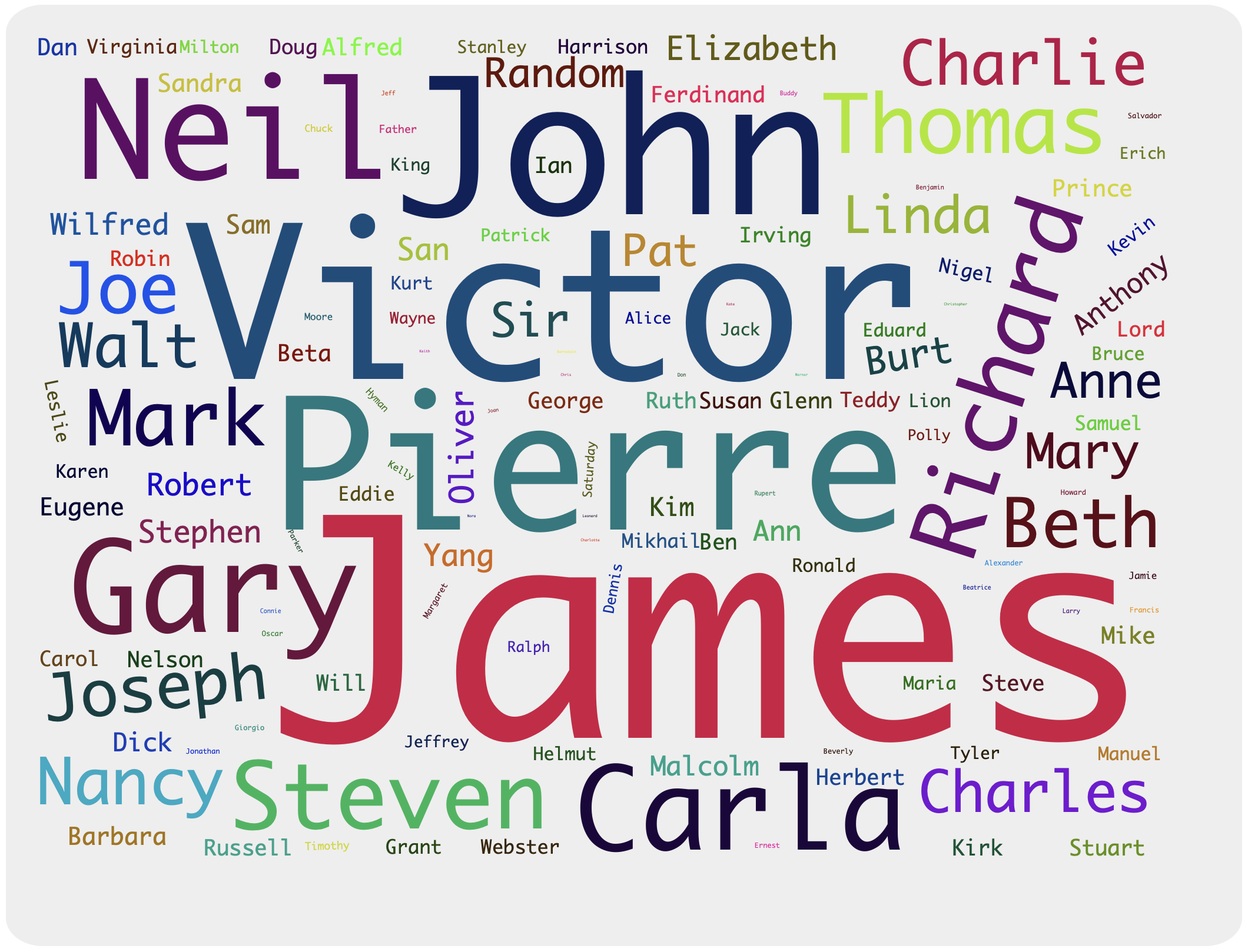}
    \caption{Semantic: Person}
    \label{fig:tv}
    \end{subfigure}    
    \caption{Examples of encoded concepts in BERT aligned with  human-defined ontologies}
    \label{fig:aligned_concept_examples}
\end{figure*}

\subsubsection{Agglomerative hierarchical clustering}

\citet{dalvi2022discovering} utilized Agglomerative hierarchical clustering to organize words. This clustering technique generates a binary tree with $N$ leaves which represent singletons (clusters of individual data points/words). Conversely, all other nodes in the tree signify clusters formed by merging the members of their respective child nodes. The merging of clusters takes place iteratively, driven by Ward's minimum variance criterion which utilizes intra-cluster variance as a measure of dissimilarity. The similarity between vector representations is evaluated using Euclidean distance. To extract a total of $K$ clusters from this hierarchical structure, the tree is cut at layer $N-K$, followed by the retrieval of nodes without parents. For instance, a cut at layer 0 yields $N$ clusters, each comprising a single point, while a cut at layer $N-1$ results in a solitary cluster containing all $N$ points. 

In terms of computational complexity, Agglomerative clustering exhibits a time complexity of $O(N^2(D + \log(N)))$ and a space complexity of $O(N^2)$ \cite{aggarwal2015data}. Here, $D$ represents the dimensionality of the data points, as the method necessitates maintaining a distance matrix that is updated throughout the algorithm's execution. The quadratic complexity constraint in $N$ confines the applicability of this method to small word datasets.

\subsubsection{Leaders Algorithm}
\label{sec:leaders}
An effective strategy for enhancing the efficiency of Agglomerative hierarchical clustering involves preprocessing the data points in order to reduce their count from $N$ to a much smaller value, denoted as $M \ll N$. The \textit{Leaders Algorithm} \cite{hartigan75} accomplishes this by making a single pass over the data in an arbitrary order. During this pass, any data point that lies within a distance of $\tau$ from a previously encountered point is classified as a \textit{follower} to the former point which becomes a \textit{leader}. Following this pass, each leader clique of connected points is condensed, often through the computation of a centroid. The resulting reduced dataset of centroids is then clustered, e.g. using Agglomerative hierarchical clustering. It is important to emphasize that the clustering outcome achieved through this method is not equivalent to directly applying Agglomerative hierarchical clustering on the original data. The results are contingent on both the arbitrary order established during the single pass and the chosen threshold value $\tau$. 

For this approach, the space complexity is $O(M^2)$, and while the clustering phase has the reduced time complexity of $O(M^2(D + \log(M)))$, the dominant factor influencing the time complexity of this algorithm is the $O(N^2)$ time complexity of the preprocessing phase.

\subsubsection{\texorpdfstring{$K$}{K}-Means clustering}

$K$-Means clustering is a widely used machine learning technique for partitioning a dataset into distinct groups or clusters. The objective of $K$-Means is to group similar data points together while maximizing the dissimilarity between different clusters. It operates by proposing a set of centroids then iteratively assigning data points to the nearest cluster centroid and  recalculating the centroids based on the newly-formed clusters. As the algorithm progresses, the clusters gradually represent coherent patterns or structures within the dataset.  This process continues until convergence when the centroids stop updating or a maximum number of iterations is reached. 

Given its susceptibility to getting trapped in local minima, the approach often incorporates random restarts to significantly enhance the exploration in the optimization space. Remarkably efficient, the $K$-Means clustering algorithm boasts a space complexity of $O(N(D + K))$, contingent on the dimensionality $D$ of the data points and the number of clusters $K$~\cite{Jin2010}. For a sequence of $I$ iterations, its time complexity stands at $O(NKDI)$~\cite{aggarwal2015data}.

\section{Assessing Quality of Concept Discovery}
\label{sec:conceptAlignment}

With the unsupervised identification of \emph{encoded concepts} by a clustering algorithm, a question arises: \emph{how can we effectively compare various clustering algorithms in relation to their ability to uncover these concepts?} We introduce a measure that evaluates the alignment of encoded concepts in light of linguistic ontologies (e.g. parts-of-speech tagging). Previous research  ~\cite{kovaleva-etal-2019-revealing,merchant-etal-2020-happens,durrani-etal-2021-transfer,durrani-etal-2022-latent} showed that higher layers of PLMs get optimized for the task that the PLM is trained for, and that tuning a PLM for any task results in its latent space being skewed towards the output classes of the target task.\footnote{We verified this through our experiment, which involved comparing latent concepts before and after fine-tuning, as detailed in Section~\ref{sec:calibrating} and Figure~\ref{fig:tunedModels}.} We use this finding to compare the concepts discovered via different algorithms, by measuring the alignment between the encoded concepts learned by a fine-tuned model to the human-defined concepts of the underlying task. Although we here employ a fine-tuned model to assess quality, it is crucial to emphasize that this is solely for evaluation purposes. The selected algorithm can subsequently be applied to any generic pre-trained language model to uncover its latent concepts as well.

Formally, consider a downstream task (e.g. POS tagging \cite{marcus-etal-1993-building}), 
for which we possess true class annotations 
 for the input data $\mathcal{D}$ (i.e.,  per-word POS labels). For each 
tag, we construct a \textit{human-defined concept} using the annotated data. For instance, $C_h(\texttt{VBD}) = \{died, smiled, explored, \dots \}$ defines a human concept comprising past-tense verbs, while $C_h (\texttt{NNS}) = \{boys, girls, rackets, \dots \}$ outlines a concept of plural nouns. Let $\mathcal{C_H} = \{C_{h_{1}}, C_{h_{2}}, \dots, C_{h_{n}} \}$ be the set of all human-defined concepts for the task, and $\mathcal{C_E} = \{C_{e_{1}}, C_{e_{2}}, \dots, C_{e_{m}} \}$ be the set of discovered encoded concepts within the latent space of the fine-tuned PLM.
We define their \textbf{$\theta$-alignment} as a function $\lambda_{\theta}(\mathcal{E}, \mathcal{H})$:

\begin{align*}
\lambda_{\theta}(\mathcal{E}, \mathcal{H}) &=\frac{1}{2} 
\frac{\sum_\mathcal{E}\alpha_{\theta}(C_e)}{|\mathcal{C_E}|} + \frac{1}{2}\frac{\sum_\mathcal{H}\kappa_{\theta}(C_h)}{|\mathcal{C_H}|} 
\text{, where}\\
  \alpha_{\theta}(C_e) &=\left\{
  \begin{array}{@{}ll@{}}
    1, & \text{if} \ \exists C_h \in \mathcal{C_H}:\frac{|C_e \cap C_h| }{|C_e|} \geq \theta  \\
    0, & \text{otherwise}
  \end{array}\right. \\
  \kappa_{\theta}(C_h) &=\left\{
  \begin{array}{@{}ll@{}}
    1, & \text{if} \ \exists C_e \in \mathcal{C_E}:\frac{|C_e \cap C_h| }{|C_e|} \geq \theta  \\
    0, & \text{otherwise}
  \end{array}\right.
  \label{eq:al2_}
\end{align*}

\noindent 
The first term 
computes the ratio of discovered concepts that are aligned (up to $\theta\in [0,1]$) to the human-defined concepts (\textit{alignment}), while the second 
measures how many unique concepts within the human-defined ontology were recovered within the latent space (\textit{coverage}). This latter term demarcates our metric from that of~\citet{dalvi2022discovering}, and we use a high threshold $\theta=0.95$  in our experiments. Figure~\ref{fig:pipeline} shows a visual representation of the two terms.

\section{Experimental Setup}

\subsection{Models and Tasks} We conducted experiments with three widely used transformer architectures: BERT-base-cased \cite{devlin-etal-2019-bert}, RoBERTa \cite{roberta}, and XLM-RoBERTa \cite{xlm-roberta}, employing their base versions (comprising 13 layers and 768 dimensions). For our investigation of clustering quality, we fine-tuned the base models on conventional tasks that encompass fundamental linguistic concepts. These tasks included 1) morphological analysis using part-of-speech tagging with the Penn TreeBank dataset \cite{marcus-etal-1993-building}; 2) syntax comprehension using CCG super tagging with the CCG TreeBank \cite{hockenmaier2006creating}; and 3) semantic tagging using the Parallel Meaning Bank dataset \cite{abzianidze-EtAl:2017:EACLshort}. Appendix \ref{sec:appendix:linguisticConcepts} and \ref{sec:tagger} provide details of these datasets and the fine-tuning setup.
We also employ two versions of Llama-2 \cite{touvron2023llama} for our experiments on large language models, specifically \texttt{Llama2-7B} and \texttt{Llama-2-7B-chat}, with an architecture of 32 layers and 10,000 embedding dimensions.

\begin{figure*}[t]
\begin{subfigure}{.49\textwidth}
    \centering
    \includegraphics[width=.88\linewidth]{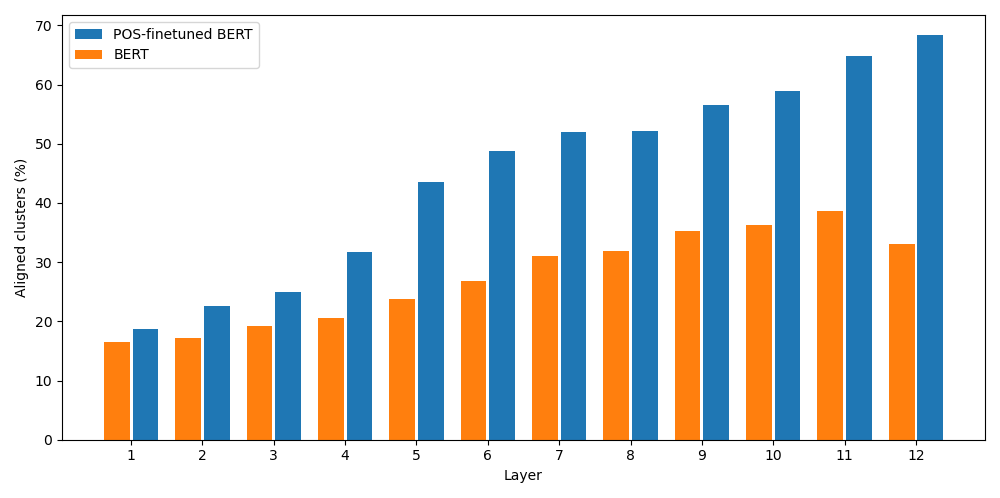}
\end{subfigure}
\begin{subfigure}{.5\textwidth}
    \centering
 \includegraphics[width=.88\linewidth]{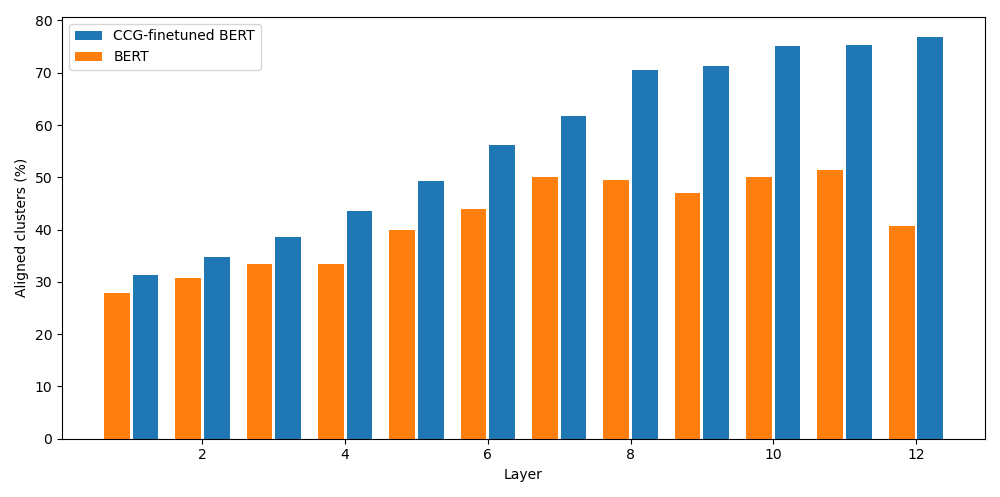}
\end{subfigure}
\caption{Alignment (percentage of discovered encoded concepts) of K-Means for POS (left) and CCG (right) in the base BERT model versus the corresponding fine-tuned models. The number of aligned concepts appreciate significantly in the higher layers of the tuned model in both cases.}
\label{fig:tunedModels}
\end{figure*}

\subsection{Calibrating latent space towards a task}
\label{sec:calibrating}
PLMs are trained towards the generic task of language modeling through next word prediction. Following the pre-training phase, the model can be fine-tuned with additional training using a more specific annotated dataset tailored to a particular task. Our approach involves fine-tuning PLMs for downstream tasks with a human ontology to align the latent space, and then aligning the discovered encoded concepts 
with the target output labels. This allows us to use the $\theta$-alignment function we described in Section \ref{sec:conceptAlignment} for clustering quality.
To quantify the extent of this transformation, we evaluate the degree of overlap between the concepts encoded within the same layer of both base models and their fine-tuned counterparts to the human concepts. As depicted in Figure~\ref{fig:tunedModels}, the number of aligned concepts increase substantially in the upper layers of the fine-tuned models compared to base models. We found this to be true for all tasks and clustering algorithms undertaken in this study.

\subsection{Clustering}
To generate data for clustering, we perform a forward-pass through the models on their respective training sets to generate contextualized feature vectors.\footnote{ We use NeuroX toolkit \cite{dalvi-etal-2023-neurox}.} Subsequently, we applied various clustering algorithms to these vectors. This process was carried out independently for each layer, resulting in the generation of $K$ clusters (i.e. encoded concepts) per layer. For our experiments, we set $K=600$ as \citet{dalvi2022discovering} found that a $K$ within the range of 600 to 1000 achieved a satisfactory balance between overly-extensive and inadequate clustering, while their exploration of other methods, such as ELbow and Silhouette, did not yield consistent outcomes. Note that $K$ here does not have to correspond to the number of classes for the target task, as each target class may further be divided into sub-classes representing different facets (e.g. Adverbs can further be split into Adverbs of time, manner, place, etc.) as found by \citet{mousi-etal-2023-llms}. 

For Agglomerative hierarchical clustering, we used the implementation of \texttt{scikit-learn}~\cite{scikit-learn} (version 0.24.2) with euclidean distance and Ward linkage criterion. For leaders, we perform binary search to find the right threshold $\tau$ in order to reduce the dataset to the desired size for a computation budget before applying Agglomerative clustering. For efficiency, the single pass that compresses the data was implemented using \texttt{Annoy} approximate neighbor library.\footnote{https://github.com/spotify/annoy} For K-Means, we also use the standard \texttt{KMeans} implementation of \texttt{scikit-learn} with sampled initial seeds and 10 restarts.

\subsection{Alignment Threshold}
\label{sec:alignmentThreshold}

We consider an encoded concept/cluster to be aligned with a human-defined concept when it exhibits an alignment of at least $95\%$ in the number of words ($\theta=0.95$), i.e. $95\%$ of the words in the cluster belong to the human-defined concept and  allowing for only a 5\% margin of noise. Nonetheless, our patterns remain consistent for lower or higher thresholds. We only consider concepts that have more than 5 unique word-types. Note that the encoded concepts are based on contextualized embedding, i.e. the same word can have different embeddings depending on the context it appears in.

\section{Results}
\label{sec:experiments}
In the following subsections, we present our comparison of the three clustering algorithms in various data and model regimes to identify the strengths and weaknesses of the underlying methods.

\subsection{Concept discovery quality}
We first compare the algorithms on exactly the same underlying dataset to answer the question \emph{how does concept discovery quality compare across clustering algorithms?}
Table~\ref{tab:EQUAL-DATA-new} shows the  metrics for three tasks and the three clustering algorithms for layer 12 embeddings of fine-tuned BERT-base-cased models. Since we are directly comparing the algorithms for quality, we use a subset of the data that all three methods are capable of processing. Note that the \textit{Leaders} variant uses \textit{Agglomerative} clustering after reducing the data to a manageable size, rendering it equivalent to Agglomerative clustering in this case.

\begin{table}[t]
	\centering
	\footnotesize
	\begin{tabular}{|c||c|c|c|c|c|}
		\hline
		\multirow{2}{*}{} &&&  \multicolumn{3}{c|}{\textbf{Layer 12}} \\ \cline{4-6}
		& \scriptsize{\textbf{Clustering}} &\scriptsize{\textbf{Size}}&\scriptsize{\textbf{Align. \%}} & \scriptsize{\textbf{Cov. \%}} &
		\scriptsize\textbf{{$\lambda_{\theta}(\mathcal{E}, \mathcal{H})$}} 
		\\ \hline  
		\multirow{2}{*}{\rotatebox[origin=c]{-90}{\textbf{pos}}}
	&{\scriptsize Agglomerative} &245K & 47.8 & \textbf{60.0} & 0.54 \\
	&{\scriptsize Leaders} &245K & 47.8 & \textbf{60.0} & 0.54 \\
	&{\scriptsize K-Means} &245K & \textbf{60.2} & \textbf{60.0} &  \textbf{0.60} \\ 
	\hline
	
	\multirow{2}{*}{\rotatebox[origin=c]{-90}{\textbf{ccg}}}
	&{\scriptsize Agglomerative} &222K & 67.2 & 22.6 & 0.45 \\
	&{\scriptsize Leaders} &222K & 67.2 & 22.6 & 0.45 \\
	&{\scriptsize K-Means} &222K & \textbf{70.5} & \textbf{23.6} & \textbf{0.47}   \\
	\hline
	
	\multirow{2}{*}{\rotatebox[origin=c]{-90}{\textbf{sem}}}
	&{\scriptsize Agglomerative} &223K &  46.8 & 50.7 & 0.49\\
	&{\scriptsize Leaders} &223K &  46.8 & 50.7 & 0.49\\
	&{\scriptsize K-Means} & 223K & \textbf{58.8} & \textbf{59.7} 
	& \textbf{0.59} \\
	\hline
\end{tabular}
\caption{Comparing aligned clusters and concepts using different clustering methods for layer 12 embeddings from dataset-fine-tuned BERT-base-cased models, while evaluating algorithm performance on identical data size.}
\label{tab:EQUAL-DATA-new}
\end{table}

\begin{figure}[t]
\centering
\includegraphics[width=0.95\linewidth]{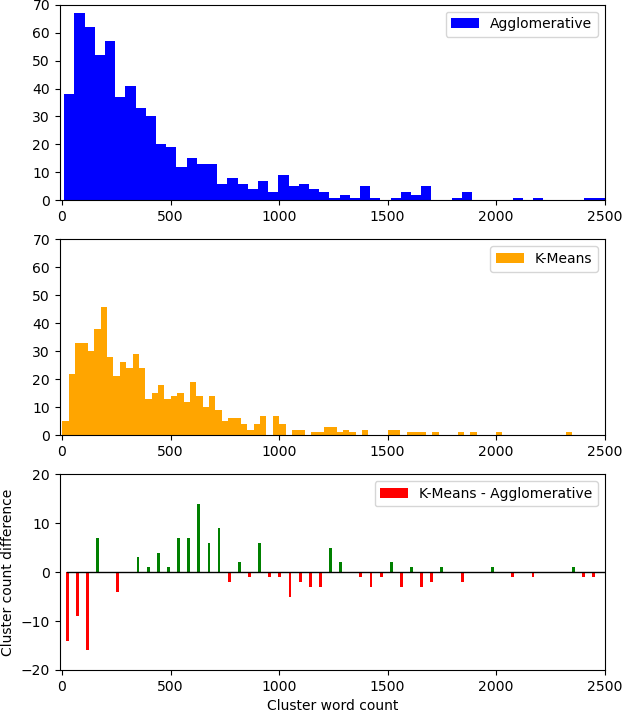}
\caption{Histogram of cluster sizes for Agglomerative hierarchical clustering and K-Means on the same data. $K$-Means shows a heavier distribution (median 319 words per cluster), while Agglomerative clustering gave more small clusters (median 275) and a longer tail.}
\label{fig:size-hist}
\end{figure}

In our results the \textit{K-Means} algorithm demonstrates a superior performance over Agglomerative clustering in alignment and coverage when using the same data. In an attempt to investigate this further, we plot the distribution of the sizes of encoded concepts  (number of words within the concept) per clustering algorithm in Figure~\ref{fig:size-hist}. As depicted in the graph, Agglomerative clustering appears to have a propensity to generate a greater number of smaller clusters than K-Means which produces more medium-sized spherical clusters. Also, it could be noted that Agglomerative clustering resulted in a longer tail in the size distribution which might relate to its sensitivity to outliers. 

\begin{table*}[t]
    \centering
    \footnotesize
    \begin{tabular}{|c||c|c|c|c|c||c|c|c||c|c|c|}
    \hline
    \multirow{2}{*}{} &&& \multicolumn{3}{c||}{\textbf{Layer 10}} & \multicolumn{3}{c||}{\textbf{Layer 11}} & \multicolumn{3}{c|}{\textbf{Layer 12}} \\ \cline{4-12}
          & \small{\textbf{Clustering}}
          &\small{\textbf{Size}}&\scriptsize{\textbf{Align. \%}} & \scriptsize{\textbf{Cov. \%}} &
          \scriptsize\textbf{{$\lambda_{\theta}(\mathcal{E}, \mathcal{H})$}} &
          \scriptsize{\textbf{Align. \%}} & 
          \scriptsize{\textbf{Cov. \%}} &
          \scriptsize\textbf{{$\lambda_{\theta}(\mathcal{E}, \mathcal{H})$}} &
          \scriptsize{\textbf{Align. \%}} & 
          \scriptsize{\textbf{Cov. \%}} &
          \scriptsize{\textbf{$\lambda_{\theta}(\mathcal{E}, \mathcal{H})$}}
          \\ \hline  
    \multirow{3}{*}{\rotatebox[origin=c]{-90}{\textbf{pos}}}
    &{\bf\scriptsize Agglomerative} &245K & 44.2 & 55.6  & 0.50 & 46.2 & 60.0 & 0.53 & 47.8 & 60.0 & 0.54 \\
    &{\bf\scriptsize Leaders} &906K & \textbf{59.2} &\textbf{64.4}  & \textbf{0.62} & 61.7 & 66.7 & 0.64 & 62.0 & 71.1 & 0.67 \\
    &{\bf\scriptsize K-Means} &906K & 58.8 & \textbf{64.4}  & \textbf{0.62} & \textbf{64.8} & \textbf{68.9} & \textbf{0.67} & \textbf{68.3} & \textbf{75.5} & \textbf{0.72} \\ 
    \hline
    
    \multirow{3}{*}{\rotatebox[origin=c]{-90}{\textbf{ccg}}}
    &{\bf\scriptsize Agglomerative} &222K & 63.3 & 19.1  & 0.41 & 65.8 & 21.0 & 0.43 & 67.2 & 22.6 & 0.45 \\
    &{\bf\scriptsize Leaders} &923K & \textbf{79.0} & 20.9 & \textbf{0.50} & \textbf{78.3} & \textbf{25.1} & \textbf{0.52} & 73.5 & 25.5 & 0.50\\
    &{\bf\scriptsize K-Means} &923K & 75.2 & \textbf{23.2} & 0.49 & 75.3  & 22.8 & 0.49 & \textbf{76.8 }& \textbf{25.9} & \textbf{0.51}   \\
    \hline
    
    \multirow{3}{*}{\rotatebox[origin=c]{-90}{\textbf{sem}}}
    &{\bf\scriptsize Agglomerative} &223K & 41.6 & 49.2  & 0.45 & 45.5 & 52.2 & 0.49 & 46.8 & 50.7 & 0.49\\
    &{\bf\scriptsize Leaders} &797K & 50.5 & 59.7  & 0.55 & 52.8 & 58.2 & 0.56 & 58.2 & 61.2 & 0.60 \\
    &{\bf\scriptsize K-Means} & 797K & \textbf{57.6}  & \textbf{62.7} & \textbf{0.60} & \textbf{61.2} & \textbf{59.7} & \textbf{0.60} & \textbf{68.0} & \textbf{67.2} 
 & \textbf{0.68} \\
    \hline
    \end{tabular}
    \caption{Evaluating clustering methods across layers [10, 11, and 12] utilizing fine-tuned BERT-base-cased models, while operating within a memory constraint of 500GB. Align \% = the percentage of encoded concepts that match the human-defined concepts within each task. Cov. \% =  percentage of distinct human-defined concepts that are acquired within the latent space. $\lambda_{\theta}(\mathcal{E}, \mathcal{H})$ corresponds to an overall score combining the alignment percentage and concept coverage of human-defined concepts within a given task.}
    \label{tab:ALL-methods-LIMIT-ALL}
\end{table*}

\subsection{Concept discovery using scaled datasets}
\label{sec:mainResult}
Given that K-Means results in higher quality concept discovery, we now ask the following question: \emph{Does scaling the underlying dataset improve concept discovery?} To answer this, we proceed to compare the three clustering algorithms 
when the algorithm operates on as large of a dataset as possible within some external constraint. In our case, we used a maximum memory capacity of 500GB.

The results are presented in Table~\ref{tab:ALL-methods-LIMIT-ALL} for layers 10, 11 and 12 of fine-tuned BERT-base-cased models.  The table illustrates the varying quantities of word representations that each clustering algorithm trains on. K-Means encountered no challenge in handling full datasets, whereas Agglomerative clustering necessitated sampling a subset of 220-250K words to operate within the memory confines. Regarding the Leaders variant of the Agglomerative clustering algorithm, an initial preprocessing of the dataset condensed  the data into a reduced collection of 220-250K centroids which are then employed for clustering, as elaborated in Section~\ref{sec:leaders}. A binary search approach was used to determine the appropriate threshold value $\tau$.

Our findings  in Table~\ref{tab:ALL-methods-LIMIT-ALL} reveal a consistent superiority of the Leaders algorithm when compared to hierarchical Agglomerative clustering that operates in a data subset. This observation suggests that the initial pass of the Leaders algorithm through the data generates a representative dataset more apt for clustering, both in terms of coverage and alignment, and that  data scaling  can reveal better structure. 

Possessing the capacity to handle the entire dataset without preprocessing, K-Means consistently outperformed the other two alternatives in terms of alignment and coverage across most scenarios. However, it does perform slightly poorly compared to Leaders in the case of CCG, specifically in layers 10 and 11.  In CCG, words are tagged with complex linguistic categories that are composed hierarchically, reflecting the grammatical relationships and syntactic structures present in the text. Therefore it is plausible that a Leaders clustering potentially benefits from its ability to capture both overarching and nuanced relationships present in the hierarchical structure of the linguistic categories. Note that CCG coverage is also limited for this reason, stemming from the intricate and diverse range of syntactic functions that define these tags.

\subsection{Computational complexity}
\label{sec:complexity}
Table~\ref{tab:complexity} lists the average computational requirements of the three clustering algorithms in our experiments. The runtime is the total of the \texttt{user} and \texttt{sys} components of the output of Linux's \texttt{time} command. For peak memory usage, we employed Python's \texttt{memory\_profiler}.\footnote{https://github.com/pythonprofilers/memory\_profiler}

\begin{table}[ht]
\footnotesize
\centering
\begin{tabular}{|c||c|c|c|c|}
\hline
& \textbf{\scriptsize{Clustering}} & \textbf{\scriptsize{Size}}& \scriptsize{\textbf{Runtime} (s)} & \scriptsize{\textbf{Memory} (GB)} \\
\hline
\multirow{3}{*}{\rotatebox[origin=c]{-90}{\textbf{pos}}}
&{Agglomerative} & 245K& 49,709& 450.38 \\
&{Leaders}& 906K& 50,967&421.43\\
&{K-Means}&906K& \textbf{32,461}&\textbf{13.59}
\\
\hline
\multirow{3}{*}{\rotatebox[origin=c]{-90}{\textbf{ccg}}}&{Agglomerative}& 223K& 39,045&371.40\\  
&{Leaders}&797K&60,599&443.94\\
&{K-Means}&797K&\textbf{37,930}&\textbf{16.17}\\
\hline
\multirow{3}{*}{\rotatebox[origin=c]{-90}{\textbf{sem}}}&{Agglomerative}& 222K&35,401&375.76\\ 
&{Leaders}&923K&49,141&394.47\\
&{K-Means}&923K&\textbf{28,991}&\textbf{13.00}\\
\hline
\end{tabular}
\caption{Runtime and memory requirements per clustering method and dataset, averaged across layers 10--12 for the results in Table~\ref{tab:ALL-methods-LIMIT-ALL}}
\label{tab:complexity}
\end{table}

As the results clearly show, K-Means demonstrates superior time efficiency and remarkably low memory requirements compared to the other two alternatives, highlighting its potential for scalability. Please note that the numbers for the Leaders algorithm solely pertain to the two-stage clustering process and do not account for the binary search procedure required to determine the appropriate threshold $\tau$ to meet the memory limitation of 500GB.

\subsection{Cross-architectural comparison}
\label{sec:crossModel}

\emph{Do our findings generalize across models?}  We reproduced the BERT-base-cased experiments comparing the clustering approaches using the final layer of RoBERTa and XLM-RoBERTa. Despite the shared foundation of transformer-based pre-trained language models, these models vary in training regime, including data, optimization functions, pre-processing, and hyperparameters, among other factors. Our findings, shown in Table~\ref{tab:MULTIMODELS}, revealed a certain trend: K-Means clustering consistently outperformed Agglomerative clustering across all scenarios. While the distinction between K-Means and the Leaders algorithm was less pronounced, K-Means remained the preferred choice due to computational requirements and potential for scalability.

\begin{table}[t]
    \centering
    \footnotesize
    \begin{tabular}{|c||c|c|c|c|c|}
    \hline
    \multirow{3}{*}{} &&&  \multicolumn{3}{c|}{\textbf{Layer 12}} \\ \cline{4-6}
          & \scriptsize{\textbf{Clustering}} &\scriptsize{\textbf{Size}}&\scriptsize{\textbf{Align. \%}} & \scriptsize{\textbf{Cov. \%}} &\scriptsize{$\lambda_{\theta}(\mathcal{E}, \mathcal{H})$} 
          \\ \hline  
    \multirow{3}{*}{\rotatebox[origin=c]{-90}{\textbf{bert}}}
    &{\scriptsize Agglomerative} &245K & 47.8 & 60.0 & 0.54 \\
    &{\scriptsize Leaders} &906K & 62.0 & 71.1 & 0.67 \\
    &{\scriptsize K-Means} &906K & \textbf{68.3} & \textbf{75.5} & \textbf{0.72} \\ 
    \hline
    \multirow{3}{*}{\rotatebox[origin=c]{-90}{\textbf{roberta}}}
    &{\scriptsize Agglomerative} &245K & 37.8 & 64.4 & 0.51 \\
    &{\scriptsize Leaders} &906K & 51.7 & 64.4 & 0.58 \\
        &{\scriptsize K-Means} &906K & \textbf{56.8} & \textbf{77.8} & \textbf{0.67}   \\
    \hline
    
    \multirow{3}{*}{\rotatebox[origin=c]{-90}{\textbf{xlm-r}}}
    &{\scriptsize Agglomerative} &245K &  44.0 & 57.8 & 0.51 \\
    &{\scriptsize Leaders} &906K & \textbf{56.5} & \textbf{64.4} & \textbf{0.60} \\
    &{\scriptsize K-Means} & 906K & 56.3 & \textbf{64.4} 
 & \textbf{0.60} \\
    \hline
    \end{tabular}
    \caption{Comparing aligned clusters and concepts using different clustering methods in BERT, RoBERTa and XLM-RoBERTa language models on POS task}
    \label{tab:MULTIMODELS}
\end{table}

\begin{figure}[t]
    \centering    \includegraphics[width=0.47\textwidth]{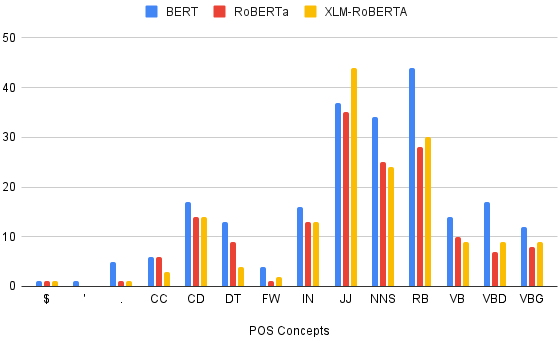}
    \caption{Number of aligned concepts for selected POS tag across different models. More results across various models and clusterings can be found in Appendix~\ref{sec:appendix:comparingArchitectures}.}
    \label{fig:comparingModels-POS-K-Means}
\end{figure}

When comparing different architectures, we observed that the concepts of BERT exhibit a stronger alignment with human-defined concepts in comparison to other models. For instance, when employing $K$-Means clustering, the percentage of concepts aligned in BERT is 68.3. 
We note that BERT shows a higher alignment of concepts across nearly all the tags compared to XLM-RoBERTa and RoBERTa, as shown in Figure~\ref{fig:comparingModels-POS-K-Means}. These findings suggest that concepts within BERT may display a greater level of redundancy compared to RoBERTa and XLM-RoBERTa. Our finding resonates with \citet{JMLR:v24:23-0074} who also found information to be more redundantly stored in BERT-base-cased model as opposed to RoBERTa and other PLMs. 

\section{Applications}
Given the larger scale that K-Means clustering offers, we present preliminary results on potential future directions in latent concept discovery that were previously not possible. Specifically, we look at concepts beyond the level of individual words, and also present results on two large language models, which have now become the foundation for many important leaps in NLP.

\subsection{Phrasal-level interpretability}
\label{sec:sequenceTasks}

While it might suffice to focus solely on words when analyzing latent spaces for sequence labeling tasks, it becomes necessary to capture extensive contextual dependencies that span across longer ranges when dealing with sentence-level tasks, e.g.  complex natural language understanding challenges like those found in the GLUE benchmark \cite{wang-etal-2018-glue}. Such tasks often revolve around extended word spans  which play a crucial role in the task, referred to as ``rationales'' \cite{deyoung-etal-2020-eraser}. For instance, in the context of sentiment classification for the following sentence, the highlighted spans specifically define a positive sentiment:
\\\\
\texttt{In this movie, ... Plots to take over the world.
\textcolor{teal}{The acting is great!} The soundtrack is run-of-the-mill,
\textcolor{teal}{but the action more than makes up for it.}}
\\

\begin{figure}[t]
    \centering
     \includegraphics[width=6.4cm]{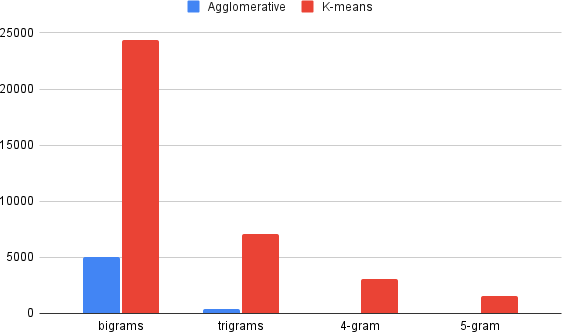}
\caption{Number of phrasal units (2- to 5-grams) discovered using hierarchical Agglomerative clustering and K-means in the latent concepts of the BERT-SST model.}
\label{fig:sst-phrasal}
\end{figure}

\begin{figure}[ht]
    \begin{subfigure}[b]{0.47\linewidth}
    \centering
    \includegraphics[width=\linewidth]{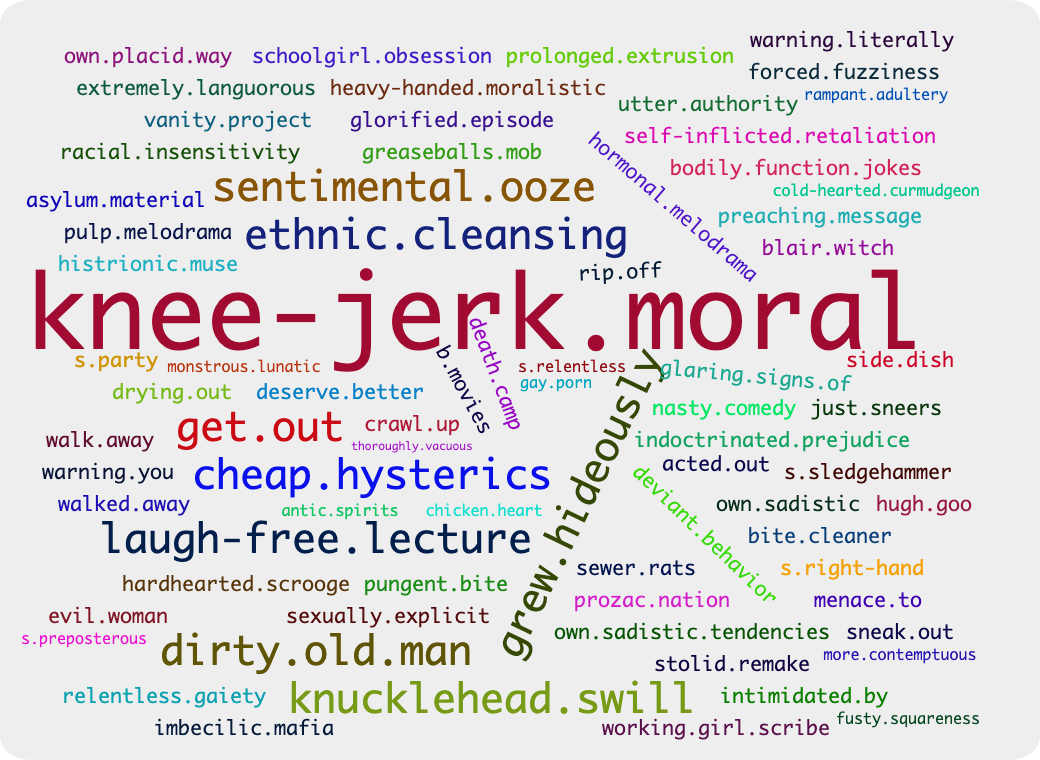}
    \caption{Negative Sentiment}
    \label{fig:negative}
    \end{subfigure}\hspace{0.15cm}
    \centering
    \begin{subfigure}[b]{0.47\linewidth}
    \centering
    \includegraphics[width=\linewidth]{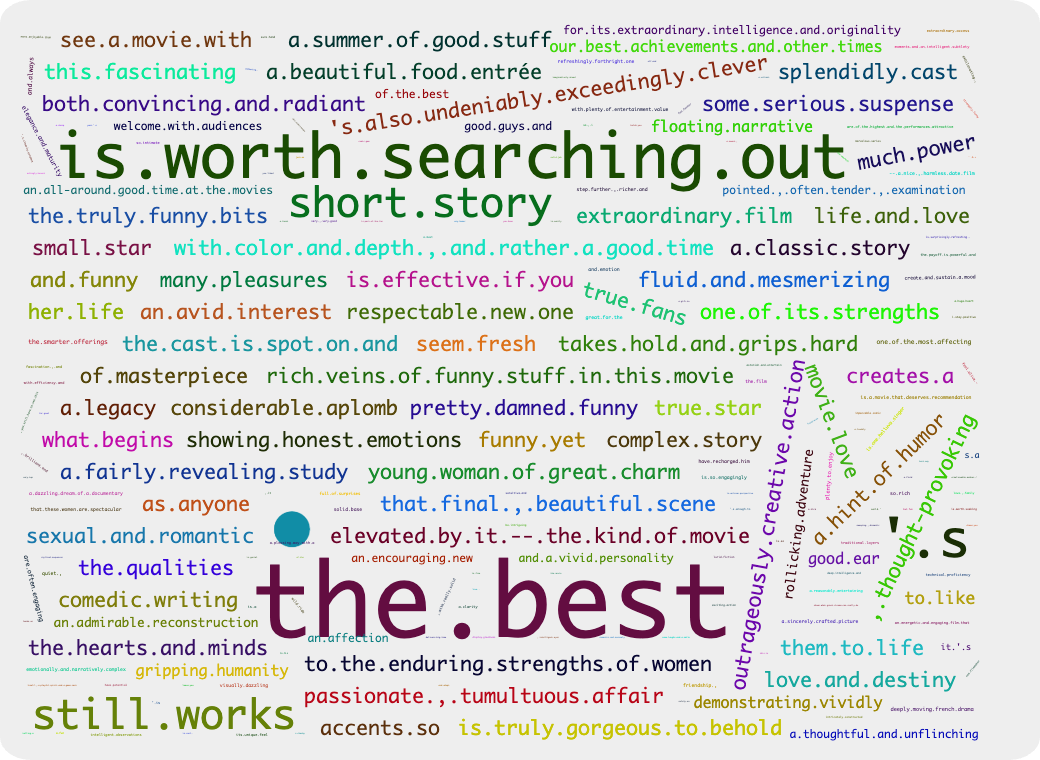}
    \caption{Positive Sentiment}
    \label{fig:positive}
    \end{subfigure}\hspace{0.15cm}
    \vspace{-5mm}
    \caption{Examples of encoded concepts in BERT-SST model. Tokens that are part of the same phrase are separated by periods.}
    \label{fig:sst-qual}
\end{figure}

We demonstrate that our scaled-up concept discovery upgrades the framework of latent concept analysis to sentence labeling tasks such as Sentiment Classification. Figure~\ref{fig:sst-phrasal} illustrates a comparison of the number of phrasal units (2 to 5-grams) found within the latent concepts of the BERT-SST model when using Agglomerative and $K$-Means clustering. The number of phrasal units discovered using $K$-Means improves significantly. Figure~\ref{fig:sst-qual} shows example polarized concepts  used in predicting negative (a) and positive (b) sentiment. We leave a detailed exploration of latent concept analysis for sentence-level tasks for the future.

\subsection{Concept discovery for LLMs}
\label{sec:llms}

\begin{figure}[t]
    \centering
    \includegraphics[width=0.37\textwidth]{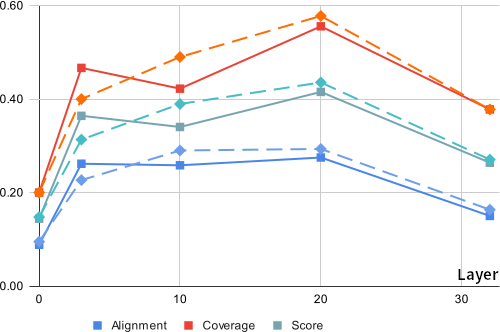}
    \caption{POS alignment, coverage and score for $K$-Means discovered concepts using layers 0, 3, 10, 20 and 32 in \texttt{Llama-2-7B} (solid) and \texttt{Llama-2-7B-chat} (dashed), showing a slightly better alignment in later layers in the chat-tuned variant.}
    \vspace{-2mm}
    \label{fig:llama2}
\end{figure}

Now we turn attention to latent concept discovery in large language models which our work facilitates. We investigate if \texttt{Llama-2-7B} and \texttt{Llama-2-7B-chat} exhibit any alignment to part-of-speech (POS) tags, similar to the models presented before. Figure~\ref{fig:llama2} shows the alignment, coverage and $0.95$-alignment score for the discovered concepts using the activations of the two LLMs. The chat-tuned version of Llama starts lower than its base sibling in alignment, but increases towards that later layers. Even though the difference is not very significant, we envisage that a suitable prompt that encourages the LLM to classify words into their respective POS tags would steer the alignment of the chat version higher. We leave this exploration of the role of the prompt for future work.

We also repeat the experiments that look at different underlying dataset sizes to verify that data scale is essential for latent concept discovery in LLMs. We recompute the alignment results for layer 20 of \texttt{Llama-2-7B} when we filter out any input word that occurs less than 5 times or more than 1000 times (similar to the filtering done for Agglomerative clustering in~\citet{dalvi2022discovering}). This results in a 50\% reduction in the input data size. Table~\ref{tab:scale} shows a big drop in alignment and coverage using the reduced data, reinforcing that a lot of the lingual structure could be removed if data is pre-processed for size.

\begin{table}[ht]
\footnotesize
    \centering
    \begin{tabular}{|c|c|c|c|}
    \hline
    & \scriptsize{\textbf{Align.}\%} & \scriptsize{\textbf{Cov.}\%} & \scriptsize{$\lambda_{\theta}(\mathcal{E}, \mathcal{H})$} \\
    \hline
         {full} & \textbf{27.50} & \textbf{55.56} & \textbf{0.41}  \\
         {filtered} & 22.17 & 28.89 & 0.25 \\
             \hline
    \end{tabular}
    \caption{Alignment, coverage and $0.95$-alignment score for layer 20 of \texttt{llama-2} when using the \textbf{full} data, and when the data is \textbf{filtered} based on word frequency. The results show the benefits of working on the full-scale of data for latent concept discovery.}
    \label{tab:scale}
\end{table}

\vspace{-2mm}
\section{Related Work}
\label{sec:relatedWork}
Clustering of representations from neural language models has been vital in a large number of studies to improve downstream NLP task and analyze language models. For instance, \citet{aharoni-goldberg-2020-unsupervised} cluster sentence embeddings to showcase the separation of domains in the embedding spaces of language models. \citet{fei-etal-2022-beyond} use clustering to improve zero-shot text classification PLMs. \citet{Gupta_Shi_Gimpel_Sachan_2022} utilized deep clustering of representations to improve zero-shot performance on Part-of-Speech and Constituency label induction. \citet{zhang-etal-2022-neural, thompson2020topic} cluster contextual embeddings to improve topic modeling of textual corpora. \citet{michael-etal-2020-asking,dalvi2022discovering,Yao-Latent} explored latent spaces of models to analyze the knowledge learned within a model. \citet{sajjad:naacl:2022,alam:2023:AAAI}  compared them to the traditional and newly-discovered human ontologies. More recently \citet{mousi-etal-2023-llms} used LLMs to annotate latent spaces learned within pre-trained LMs. Our work addresses an important challenge underpinning these and similar works: the high computational cost of clustering in large embedding spaces.

\section{Conclusion}
\label{sec:conclusion}

Concluding this study, our exploration into uncovering latent concepts within the embedding space of pre-trained language models represents just the initial phase towards comprehending these models and establishing trust in their functionality. Our findings underscore the effectiveness of employing clustering of contextualized representations to unveil meaningful concepts that resonate with human comprehension. Furthermore, we highlight the viability of utilizing K-Means algorithm to handle expansive datasets, thereby facilitating the analysis of larger models and more intricate concepts.

Moving forward, there remains a compelling avenue for delving deeper into the interpretability of language models beyond the granularity of individual words. This involves unraveling the representation and utilization of complex linguistic constructs for inference purposes. Also, our exploration of the Leaders algorithm could potentially expand beyond our use with Agglomerative clustering. For instance, an intriguing avenue for future research includes enhancing the scalability of K-Means, pushing its limits to accommodate even more extensive datasets.

\section*{Limitations}

The results presented in the paper mainly revolve around the proposed metric that combines alignment and coverage of human-defined ontologies. 
Although this metric serves as a reliable proxy for assessing the quality of clustering, it may not explicitly capture other dimensions of clustering that are equally important but not accounted for. Furthermore, we have limited our experimentation to three clustering algorithms among the many available, and it is conceivable that there are other algorithms that may result in even better quality at similar or lower computational cost than K-Means. Finally, our applications section mainly presents high level results without very deep exploration of the underlying hyperparameters, as these could be complete works in their own regard.

\bibliography{eacl24,custom}

\section*{Appendix}
\label{sec:appendix}
\appendix
\section{Linguistic Concepts}
\label{sec:appendix:linguisticConcepts}

We used parts-of-speech tags (48 concepts) using Penn Treebank data \cite{marcus-etal-1993-building}, semantic tags (73 concepts) \cite{abzianidze-EtAl:2017:EACLshort}, and CCG super tags (1272 concepts). Please see all the concepts below in Tables~\ref{tab:penn_treebank_pos_tags} and \ref{tab:sem-tags}. This provides a good coverage of linguistic concepts including morphology, syntax and semantics.

\begin{table}[!tbh]
\centering
\scalebox{0.75}{
\setlength{\tabcolsep}{2.0pt}
\begin{tabular}{@{}lll@{}}
\toprule
\textbf{\#} & \textbf{Tag} & \textbf{Description} \\ \midrule
1 & CC & Coordinating conjunction \\
2 & CD & Cardinal number \\
3 & DT & Determiner \\
4 & EX & Existential there \\
5 & FW & Foreign word \\
6 & IN & Preposition or subordinating conjunction \\
7 & JJ & Adjective \\
8 & JJR & Adjective, comparative \\
9 & JJS & Adjective, superlative \\
10 & LS & List item marker \\
11 & MD & Modal \\
12 & NN & Noun, singular or mass \\
13 & NNS & Noun, plural \\
14 & NNP & Proper noun, singular \\
15 & NNPS & Proper noun, plural \\
16 & PDT & Predeterminer \\
17 & POS & Possessive ending \\
18 & PRP & Personal pronoun \\
19 & PRP\$ & Possessive pronoun \\
20 & RB & Adverb \\
21 & RBR & Adverb, comparative \\
22 & RBS & Adverb, superlative \\
23 & RP & Particle \\
24 & SYM & Symbol \\
25 & TO & to \\
26 & UH & Interjection \\
27 & VB & Verb, base form \\
28 & VBD & Verb, past tense \\
29 & VBG & Verb, gerund or present participle \\
30 & VBN & Verb, past participle \\
31 & VBP & Verb, non-3rd person singular present \\
32 & VBZ & Verb, 3rd person singular present \\
33 & WDT & Wh-determiner \\
34 & WP & Wh-pronoun \\
35 & WP\$ & Possessive wh-pronoun \\
36 & WRB & Wh-adverb \\
37 & \# & Pound sign \\
38 & \$ & Dollar sign \\
39 & . & Sentence-final punctuation \\
40 & , & Comma \\
41 & : & Colon, semi-colon \\
42 & ( & Left bracket character \\
43 & ) & Right bracket character \\
44 & " & Straight double quote \\
45 & ' & Left open single quote \\
46 & " & Left open double quote \\
47 & ' & Right close single quote \\
48 & " & Right close double quote \\ \bottomrule
\end{tabular}%
}
\caption{Penn Treebank POS tags.}
\label{tab:penn_treebank_pos_tags}
\end{table}
 
\begin{table*}[!tbh]
\centering
\scalebox{0.75}{
\setlength{\tabcolsep}{2.0pt}
\begin{tabular}{@{}llll@{}}
\toprule
\textbf{ANA   (anaphoric)} &  & \multicolumn{2}{l}{\textbf{MOD   (modality)}} \\ \midrule
PRO & anaphoric \& deictic pronouns: he, she, I, him & NOT & negation: not, no, neither, without \\
DEF & definite: the, loIT, derDE & NEC & necessity: must, should, have to \\
HAS & possessive pronoun: my, her & POS & possibility: might, could, perhaps, alleged, can \\
REF & reflexive \& reciprocal pron.: herself, each other & \multicolumn{2}{l}{\textbf{DSC (discourse)}} \\
EMP & emphasizing pronouns: himself & SUB & subordinate relations: that, while, because \\
\textbf{ACT (speech act)} &  & COO & coordinate relations: so, \{,\}, \{;\}, and \\
GRE & greeting \& parting: hi, bye & APP & appositional relations: \{,\}, which, \{(\}, — \\
ITJ & interjections, exclamations: alas, ah & BUT & contrast: but, yet \\
HES & hesitation: err & \multicolumn{2}{l}{\textbf{NAM (named entity)}} \\
QUE & interrogative: who, which, ? & PER & person: Axl Rose, Sherlock Holmes \\
\textbf{ATT (attribute)} &  & GPE & geo-political entity: Paris, Japan \\
QUC\* & concrete quantity: two, six million, twice & GPO\* & geo-political origin: Parisian, French \\
QUV\* & vague quantity: millions, many, enough & GEO & geographical location: Alps, Nile \\
COL\* & colour: red, crimson, light blue, chestnut brown & ORG & organization: IKEA, EU \\
IST & intersective: open, vegetarian, quickly & ART & artifact: iOS 7 \\
SST & subsective: skillful surgeon, tall kid & HAP & happening: Eurovision 2017 \\
PRI & privative: former, fake & UOM & unit of measurement: meter, \$, \%, degree Celsius \\
DEG\* & degree: 2 meters tall, 20 years old & CTC\* & contact information: 112, info@mail.com \\
INT & intensifier: very, much, too, rather & URL & URL: \url{http://pmb.let.rug.nl} \\
REL & relation: in, on, 's, of, after & LIT\* & literal use of names: his name is John \\
SCO & score: 3-0, grade A & NTH\* & other names: table 1a, equation (1) \\
\multicolumn{2}{l}{\textbf{COM   (comparative)}} & \multicolumn{2}{l}{\textbf{EVE (events)}} \\
EQU & equative: as tall as John, whales are mammals & EXS & untensed simple: to walk, is eaten, destruction \\
MOR & comparative positive: better, more & ENS & present simple: we walk, he walks \\
LES & comparative negative: less, worse & EPS & past simple: ate, went \\
TOP & superlative positive: most, mostly & EXG & untensed progressive: is running \\
BOT & superlative negative: worst, least & EXT & untensed perfect: has eaten \\
ORD & ordinal: 1st, 3rd, third & \multicolumn{2}{l}{\textbf{TNS (tense \& aspect)}} \\
\multicolumn{2}{l}{\textbf{UNE   (unnamed entity)}} & NOW & present tense: is skiing, do ski, has skied, now \\
CON & concept: dog, person & PST & past tense: was baked, had gone, did go \\
ROL & role: student, brother, prof., victim & FUT & future tense: will, shall \\
GRP\* & group: John \{,\} Mary and Sam gathered, a group of people & PRG\* & progressive: has been being treated, aan hetNL \\
\multicolumn{2}{l}{\textbf{DXS (deixis)}} & PFT\* & perfect: has been going/done \\
DXP\* & place deixis: here, this, above & \multicolumn{2}{l}{\textbf{TIM (temporal entity)}} \\
DXT\* & temporal deixis: just, later, tomorrow & DAT\* & full date: 27.04.2017, 27/04/17 \\
DXD\* & discourse deixis: latter, former, above & DOM & day of month: 27th December \\
\multicolumn{2}{l}{\textbf{LOG (logical)}} & YOC & year of century: 2017 \\
ALT & alternative \& repetitions: another, different, again & DOW & day of week: Thursday \\
XCL & exclusive: only, just & MOY & month of year: April \\
NIL & empty semantics: \{.\}, to, of & DEC & decade: 80s, 1990s \\
DIS & disjunction \& exist. quantif.: a, some, any, or & CLO & clocktime: 8:45 pm, 10 o'clock, noon \\
IMP & implication: if, when, unless &  &  \\
AND & \multicolumn{2}{l}{conjunction \& univ. quantif.:   every, and, who, any} &  \\ \bottomrule
\end{tabular}%
}
\caption{Semantic tags.}
\label{tab:sem-tags}
\end{table*}

\begin{table}[ht]									
\centering					
 \footnotesize
\scalebox{1.0}{
\setlength{\tabcolsep}{2.5pt}
    \begin{tabular}{l|ccc|l|c}							
    \toprule									
Task    & Train & Dev & Test & Tags & F1\\		
\midrule
    POS & 36.5K & 1802 & 1963 & 48 & 96.81\\
    CCG &  39.1K & 1908 & 2404 & 1272 & 95.24\\
    SEM & 36.9K & 5301 & 10600 & 73 & 96.32 \\
    \bottomrule
    \end{tabular}
    }
    \caption{Statistics of the datasets used in the experiments, the number of concepts (tags) for each task and the performance of the fine-tuned models}
\label{tab:dataStats}						
\end{table}

\section{Sequence Tagger}
\label{sec:tagger}

We performed fine-tuning on the pre-trained language models for each of the three tasks (POS, CCG and SEM tagging) used in our analysis. This entails adjusting the latent space of the models  towards the output classes, allowing us to evaluate clustering algorithms through the alignment function outlined in Section \ref{sec:conceptAlignment}. We used standard splits for training, development and test data for the  tasks. The splits to preprocess the data were  released with \citet{liu-etal-2019-linguistic} on github.\footnote{\url{https://github.com/nelson-liu/contextual-repr-analysis}} See Table \ref{tab:dataStats} for statistics and classifier accuracy for BERT-base-cased model. Appendix \ref{sec:appendix:linguisticConcepts} presents a comprehensive list of human-defined concepts within these ontologies.

\section{Comparing Architectures}
\label{sec:appendix:comparingArchitectures}

\begin{figure}[t]
    \centering
    \includegraphics[width=7.5cm]{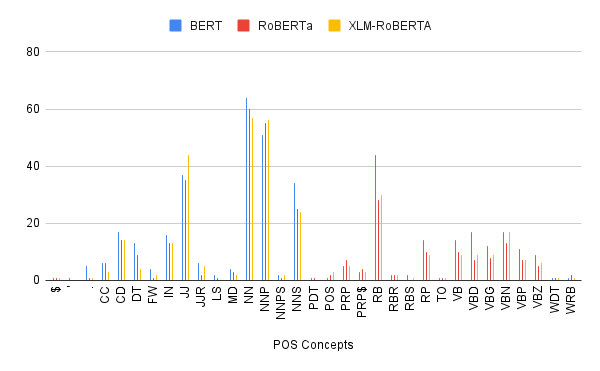}
    \includegraphics[width=7.5cm]{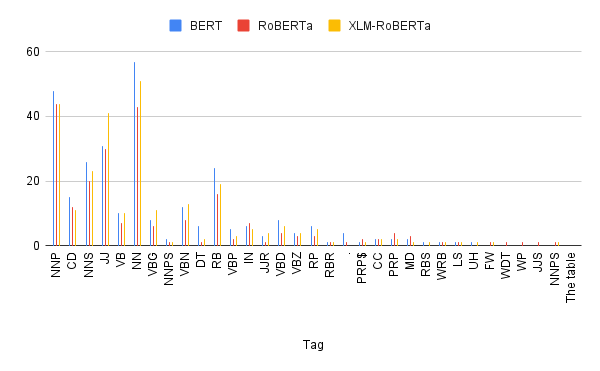}
    \caption{Number of aligned concepts per POS tag across different models}
    \label{fig:comparingModels-Full}
\end{figure}

We reproduced experiments comparing various clustering approaches using the final layer of different models. In particular, we examined the outcomes from BERT, RoBERTa, and XLM-RoBERTa. In Figure~\ref{fig:comparingModels-Full} we plot number of encoded concepts per POS tag for encoded concepts obtained via Agglomerative and K-Mean Clustering. Concepts in BERT are redundantly stored. The patterns hold consistently.

\end{document}